# Using a Catenary Trajectory to Reduce Wellbore Friction in Horizontal Extended Reach Drilling

Vu Nguyen

University of Louisiana at Lafayette

Corresponding email: vu.nguyen1@louisiana.edu

**Abstract**

Wellbore friction is one of the biggest concerns when drilling due to its relation to the total cost. The catenary concept was introduced to reduce wellbore friction, but it requires detailed analyses. This project would fill this gap. A catenary shape is simply the natural shape of a rope, chain, or drill string. The drill string will then hang freely inside the wellbore. Perfectly, there should be no contact between the hole and the string, and thus no friction. Torque and drag should be minimized this way. A case study is introduced to examine the outcome between Catenary Trajectory Design and traditional 2D Arc design. The calculation procedure of Catenary Trajectory and 2D Arc Design can be found in an MS Excel spreadsheet which is easy to use and reliable for designing catenary well trajectories for extended-reach wells.

**Introduction**

Climate change is one of the most concerning topics in our world today. The Carbon Capture Utilization Storage is not the only solution to combat this issue (Nguyen et al., 2021; Olayiwola et al., 2023). There are many new existing technologies in the campaign to reduce greenhouse gas emissions, such as hydrogen, renewable energy, etc. While increasing the efficiency of existing oil wells by enhanced oil recovery ( Carrera et al., 2023) and maintaining the condition of the wells (Olayiwola et al., 2022a; Nguyen et al., 2022; Mahmood et al., 2023; Olayiwola et al., 2022b), extended reach drilling activities also play an essential role in making energy transition become smoother by meeting oil &gas global demand. One of the innovative methods in drilling technology is using a Catenary Trajectory.

The extended reach drilling has been studied by many researchers ( El Sabeh et al., 2023; Huang and Guo, 2022; Hussain et al., 2021). El Sabeh et al. (2023) reviewed the issues of extended reach drilling, including high torque and drag, poor hole cleaning, bottom hole assembly design, and hydraulic and Equivalent Circulating Density (ECD) management, while Huang and Gao (2022) analyzed the drilling difficulty and still emphasized that high friction and drag is one of the biggest challenges. Hussain et al. (2021) introduced the help of technologies to intervene in extended-reach drilling activities, such as using mapping while drilling, Magnetic Resonance while drilling service, and advanced polyglycol water-based mud systems.

Well reach has significantly increased in the past decade. In the 1980s, a 3 km step out was common. This increased in the 1990s, and one targeted 10 km as a likely step out. Ryan et. al (1995) conducted a review of technology in 1995 and concluded that 10 km step out is feasible. This was convinced a few years later on Wytch Farm (2000), where it was also justified that shallow extended-reach wells could be drilled.

Mason and Judzis (1998) also thoroughly review and recognize technological constraints for reaching up to 18 km. Although such reach will require fit-for-purpose equipment, it would be possible in a few years. The paper will primarily inspect the potential for a catenary build section to achieve these goals.

The catenary trajectory design was first recommended to the oil industry by McClendon and Anders (1985). Several attempts have been tested, but the method has had limited application. Aadnoy and Andersen (1998) and (2001) developed the general catenary solution, which is suitable for any inclination. Ma et al. (1998) introduced field cases for catenary wells drilled in China. Torque and drag are always a big concern for all types of extended-reach well.

Planeix and Fox (1979) published a method to calculate the final angle and direction of a well that turns and builds an angle to reach an expected target from a known surface location. Unluckily, their method contains only one turn rate and must decide the turn-end point through calculation. McMilian (1981) implemented a technique for deciding the lead angle to offset the bit walk effects. Maidla and Sampaio (1989) calculated the bit walk rate and the lead angle based on a rock-bit interaction. The model was validated with the data from 15 directional wells, and the results present that bit-walk prediction was good for most of the well trajectories, but additional field examples are needed to further justify the method. The above discussion shows that 3D bit-walk paths will not be possible at a fundamental level. A set of characteristic parameter values would determine the shape of a well profile. The crucial point is that the best solution has to include the well friction as the deliberated index. With the catenary shape, a drill string will have no contact with the borehole wall. Therefore, the drill string will tend to stand off the borehole wall, and the drag and torque can be minimized.

This paper presents a 2D model of catenary trajectory design, which is easy and convenient to use. The solutions of the catenary design are in closed form and do not demand thorough numerical estimations. A traditional arc well design is also included to compare the hook load with the catenary trajectory well design.

**Mathematical model (Method description)**

In a 2D Cartesian Coordinate system, a simple Catenary Curve can be given as a hyperbolic function.

$$y = a\cosh\left(\frac{x}{a}\right) \quad (1)$$

Where a is the intercept of the catenary curve with the y axis

The vertical displacement V in catenary section can be written in term of horizontal displacement S as the following equation:

$$V = V_{end} - \left\{\frac{a}{2}\left[e^{\frac{S-S_{end}}{a}} + e^{-\frac{S-S_{end}}{a}}\right] - a\right\} \quad (2)$$

Where $V_{end}$ is the total vertical displacement in the catenary section

$S_{end}$ is the total horizontal displacement in the catenary section.

The curvature of catenary section, C, can be expressed as:

$$C = \frac{-\frac{1}{2a}\left[e^{\frac{S-S_{end}}{a}} + e^{-\frac{S-S_{end}}{a}}\right]}{\left(1 + \frac{1}{4}\left[e^{\frac{S-S_{end}}{a}} - e^{-\frac{S-S_{end}}{a}}\right]^2\right)^{\frac{3}{2}}} \quad (3)$$

The radius of curvature at top of catenary section, R, can be estimated by:

$$R = \frac{1}{C} \quad (4)$$

The build rate of the arc section or at the top of catenary section in degree per 100 feet, B, is:

$$B = 5730C \quad (5)$$

The inclination angle θ can be calculated as:

$$\theta = \frac{\pi}{2} - \alpha = \frac{\pi}{2} - \arctan\frac{dV}{dS} \quad (6)$$

Where α is slope angle

In a 2D Cartesian Coordinate system, the Kick of Point (KOP) for Arc design can be decided by:

$$VD_{KOP} = VD_{target\ base} - \frac{5730}{B_{min}}\left(\sin(I_f) - \sin(I_i)\right) \quad (7)$$

Where $I_f$ is inclination angle at target base; $I_i$ is inclination angle at KOP

For upper curve section:

$$VD_2 = VD_1 + \frac{5730}{B_{high}}[\sin(I_2) - \sin(I_1)] \quad (8)$$

$$HD_2 = HD_1 + \frac{5730}{B_{high}}[\cos(I_1) - \cos(I_2)] \quad (9)$$

$$MD_2 = MD_1 + 100\left[\frac{I_2 - I_1}{B_{high}}\right] \quad (10)$$

Where VD is vertical displacement; HD is horizontal displacement; MD is measured depth.

For tangent section:

$$VD_2 = VD_1 + \Delta MD \cos(I_{tan}) \quad (11)$$

$$HD_2 = HD_1 + \Delta MD \sin(I_{tan}) \quad (12)$$

$$MD_2 = MD_1 + \Delta MD \quad (13)$$

For lower curve section:

$$VD_2 = VD_1 + \frac{5730}{B_{low}}[\sin(I_2) - \sin(I_1)] \quad (14)$$

$$HD_2 = HD_1 + \frac{5730}{B_{low}}[\cos(I_1) - \cos(I_2)] \quad (15)$$

$$MD_2 = MD_1 + 100\left[\frac{I_2 - I_1}{B_{low}}\right] \quad (16)$$

**Results**

**A Case Study**

The data used for designing arc trajectory and catenary trajectory is presented in Table 1 and 2, respectively.

**Table 1**: Data for arc trajectory design

| Description | Value | Unit |
|---|---|---|
| Target Depth | 12500 | ft |
| Azimuth | 45 | degree |
| Build Rate | 0.691 | degree/100f t |
| Horizontal wellbore length | 7500 | ft |
| Inclination angle at target base | 90 | degree |

**Table 2**: Data for catenary trajectory design

| Description | Value | Unit |
|---|---|---|
| Total measured depth | 24000 | ft |
| Target depth | 12500 | ft |
| Vertical displacement in the catenary section $V_{end}$ | 2000 | ft |
| Horizontal displacement in the catenary section $S_{end}$ | 4000 | ft |
| Azimuth | 45 | degree |

The design for catenary trajectory includes three parts: Arc section, Catenary section, and slant section. The Kick of Point (KOP) is required before starting for Arc, Catenary, and Slant section design. The arc length is calculated based on the radius of curvature and inclination angle at the top of the catenary section. The most complicated design lies in the catenary section where inclination angle, curvature, and radius are in Eqn.6, Eqn. 3, and Eqn.4, respectively, are required to design the catenary section. The Vertical, Horizontal, North, and East Displacement calculations, which can be found in the Excel Spreadsheet, are also computed to plot the trajectory. The plot of Vertical Displacement versus Horizontal Displacement and North Displacement versus East Displacement are presented in Figures 1 and 2, respectively.

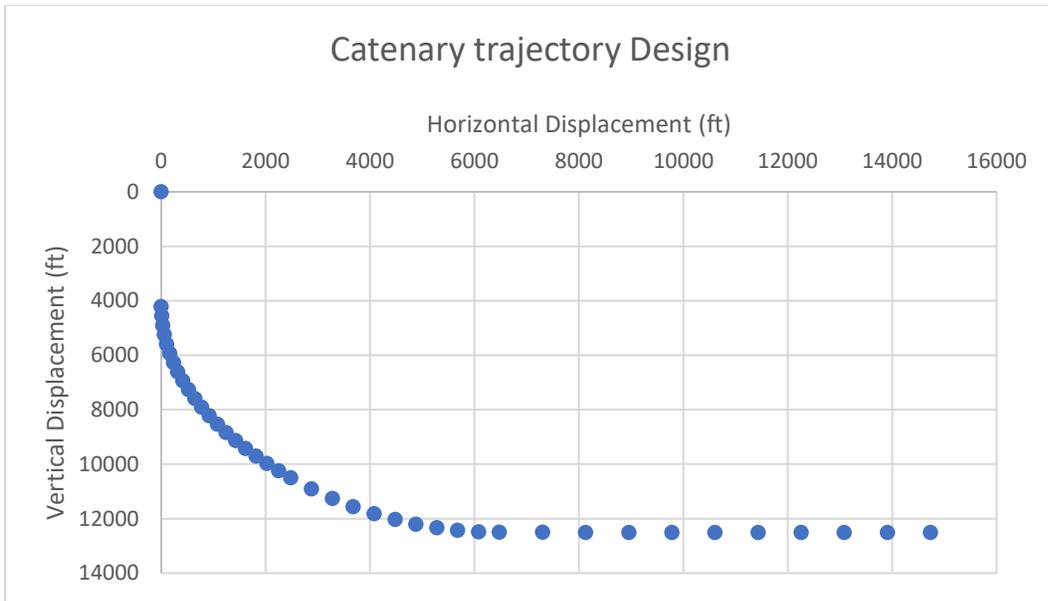

**Figure 1**: Vertical Displacement compared to Horizontal Displacement

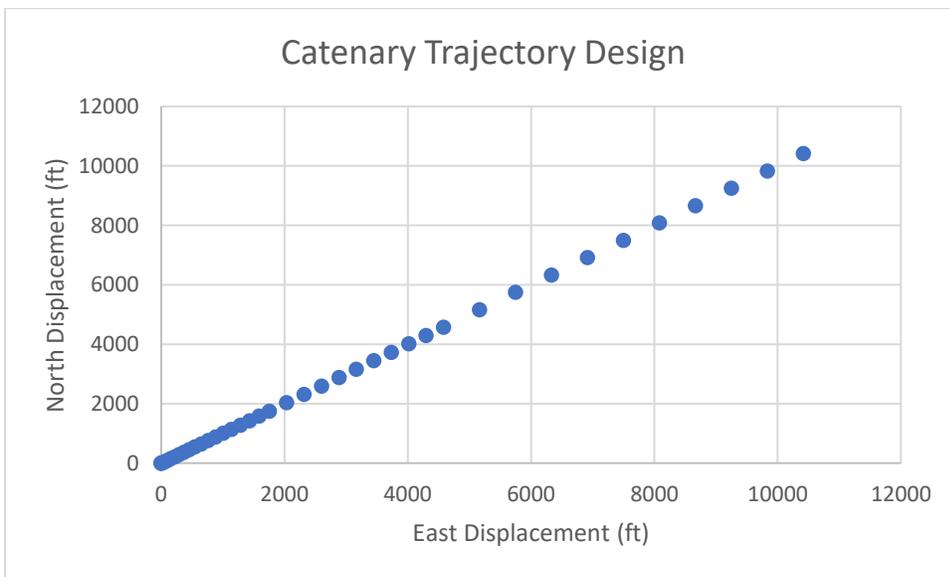

**Figure 2**: North Displacement compared to East Displacement

Similarly, the Vertical Displacement versus Horizontal Displacement and North Displacement versus East Displacement are presented in Figure 3 and 4, respectively.

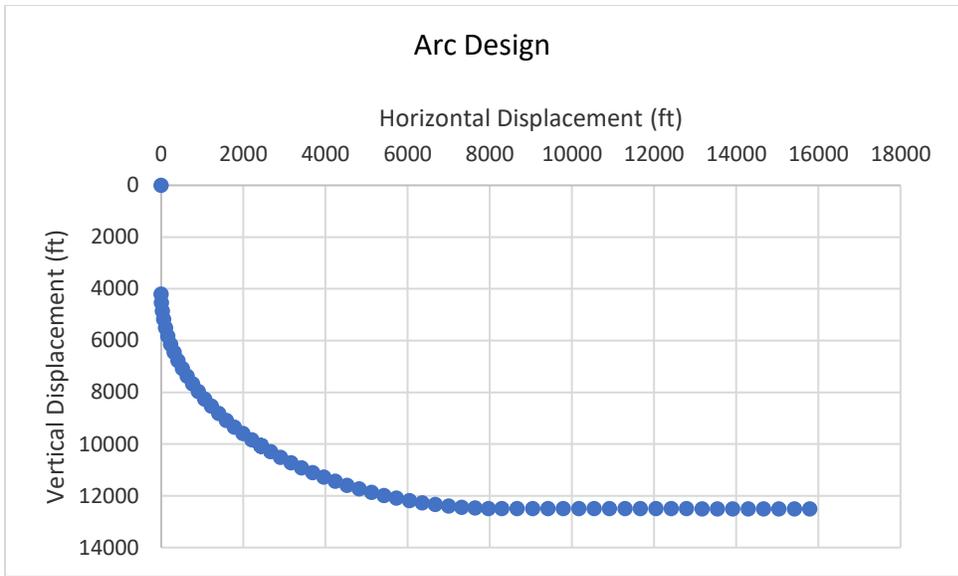

**Figure 3**: Vertical Displacement compared to Horizontal Displacement

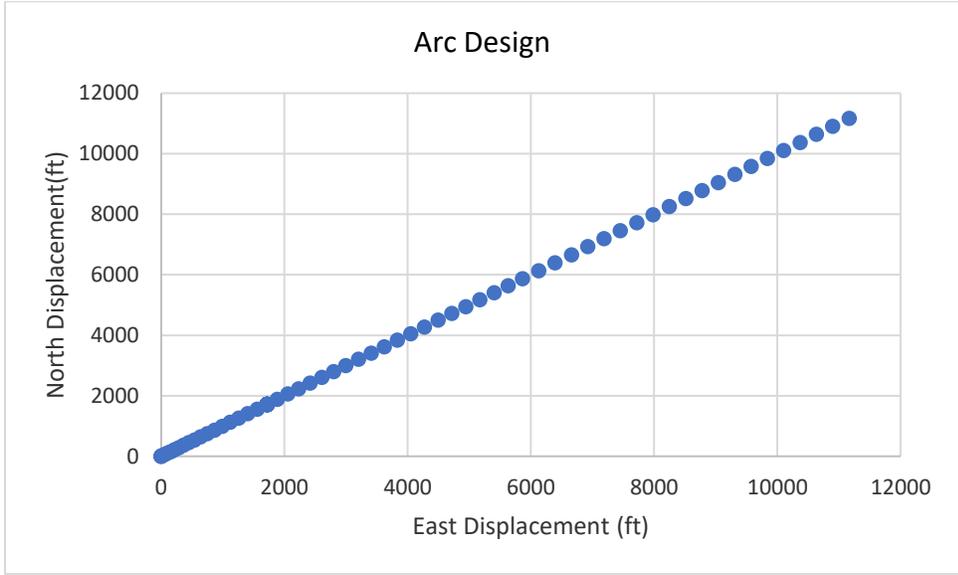

**Figure 4**: North Displacement compared to East Displacement

To have a better observation, the data of two design methods is plotted together to compare as shown in Figure 5 and 6.

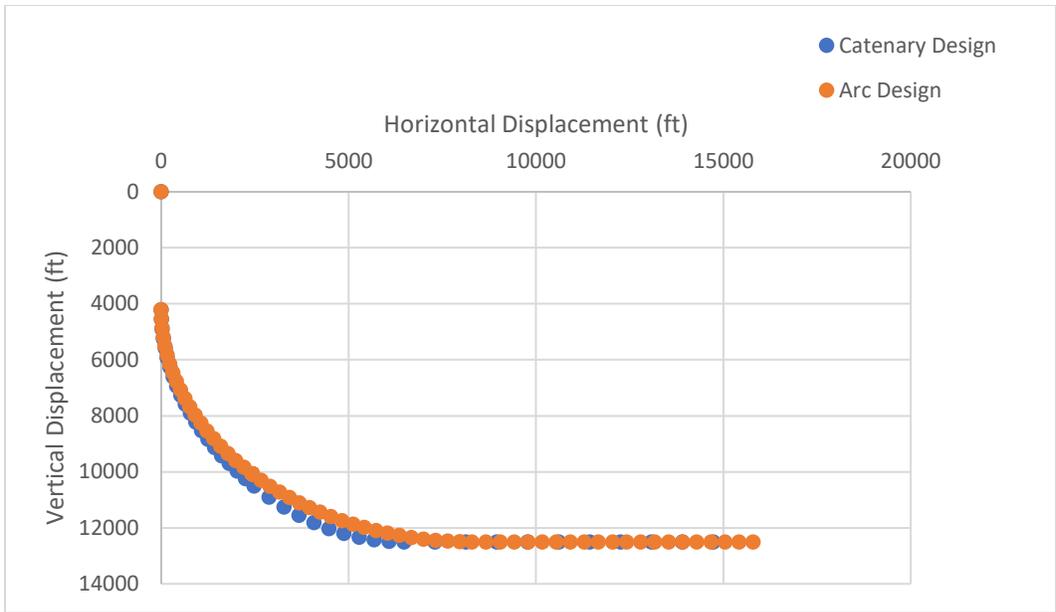

**Figure 5**: Trajectory profile comparison between two methods for Vertical Displacement versus Horizontal Displacement

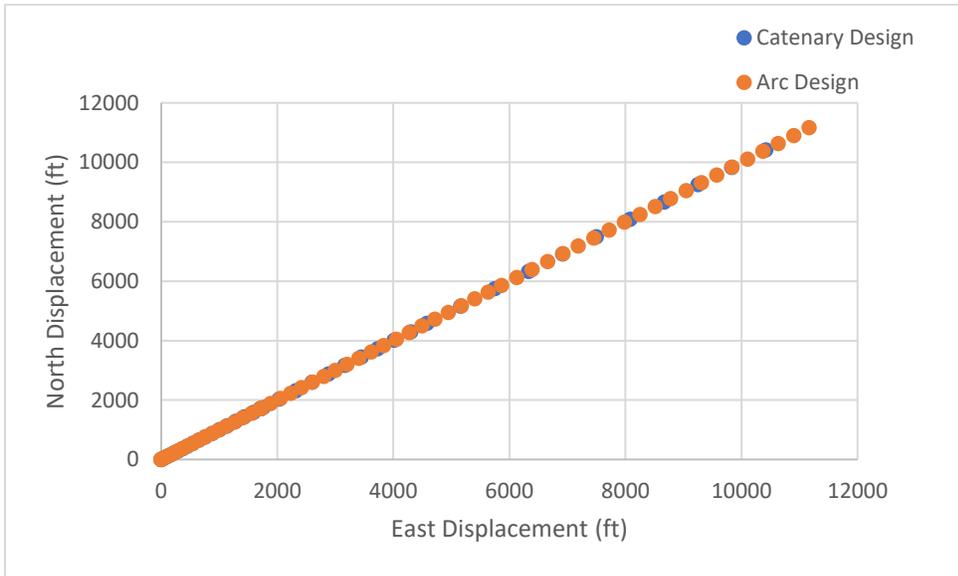

**Figure 6**: Trajectory profile comparison between two methods for North Displacement versus East Displacement

Then the total hook load (tension at surface T) analysis between two methods are performed to investigate the reduction of wellbore friction in the catenary design.

**For 2D Arc design:**

The axial compressive force at the heel can be expressed by the following equation:

$$F_{\pi/2} = \mu * W_h \quad (17)$$

The axial force in the curve section can be computed as:

$$F_O = F_{\frac{\pi}{2}} + w_C R(\mu + 1) \quad (18)$$

The Total hook load in Figure 7 is the combination of axial force in the curve section and Vertical force Wv generated by pipe weight:

$$T = F_o + W_v \quad (19)$$

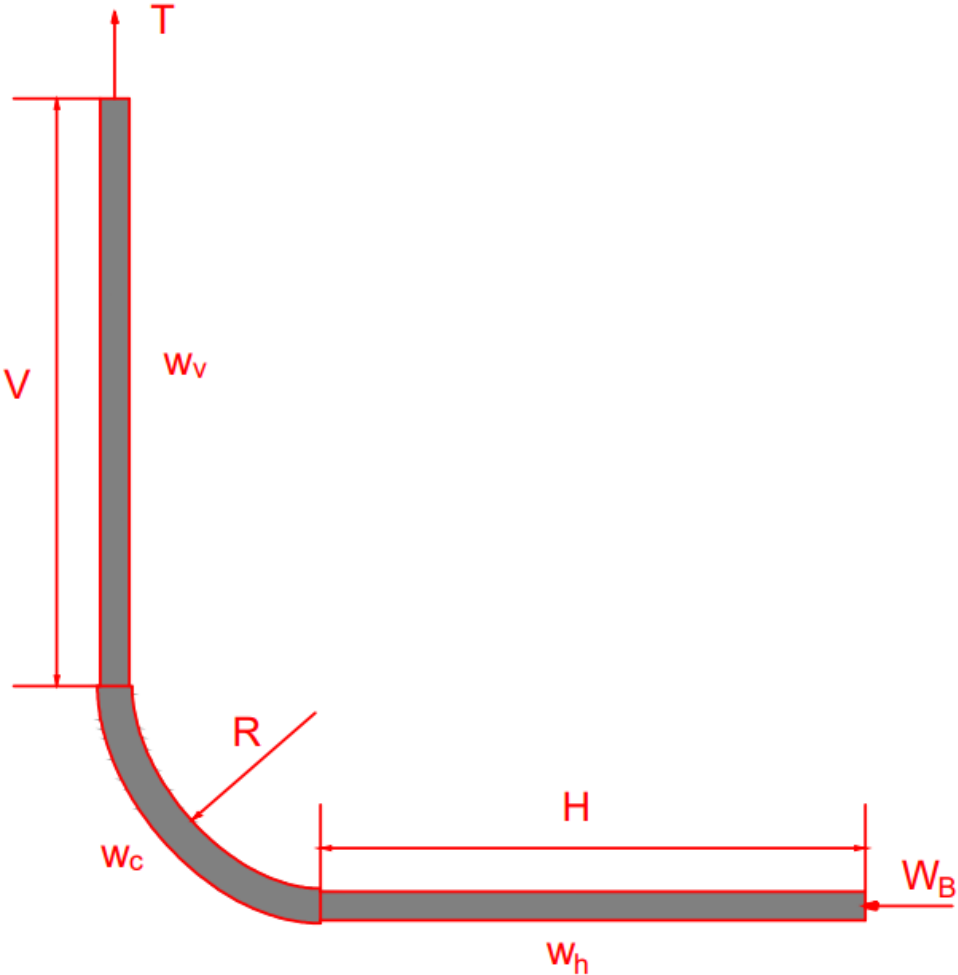

**Figure 7**: Arc design diagram

**For Catenary Trajectory Design**:

The axial compressive force at the heel has a similar expression with Arc design:

$$F_{\pi/2} = \mu * w_h \quad (20)$$

The influence of catenary shape on axial compressive force at the heel can be expressed as F$_{ct}$:

$$F_{ct} = F_{\frac{\pi}{2}} / \sin(I) \quad (21)$$

The axial force in the catenary can be computed as:

$$F_O = F_{ct} + w_c R[\sin(I) + \mu(1 - \cos(I))] \quad (22)$$

The total hook load is the combination of axial force in the curve section and Vertical force Wv generated by pipe weight:

$$T = F_o + W_V \quad (23)$$

**Hook load Analysis for the case study**

The data for pipe weight and friction coefficient is given below in table 3 to calculate the hook load.

Table 3: The pipe weight and friction values for each section

| Parameter | Vertical Section | Curve Section | Horizontal Section |
|---|---|---|---|
| Pipe weight (lbf/ft) | 19.5 | 91.69 | 16.25 |
| Friction Coefficient |  | 0.35 | 2 |

The calculation is summarized in table 4.

Table 4: Calculation of hook load

| Data Input | Arc Design | Catenary design | Unit |
|---|---|---|---|
| Horizontal friction coefficient, $\mu$ | 2 | 2 |  |
| Horizontal pipe weight section, $w_h$ | 16.25 | 16.25 | lbf/ft |
| Horizontal length | 7500 | 7430 | ft |
| Curve friction coefficient, $\mu_c$ | 0.35 | 0.35 |  |
| Curve pipe weight section, $w_c$ | 91.69 | 91.69 |  |
| Radius of curvature, R | 8,292 | 9,228 | lbf/ft |
| Inclination angle, I |  | 42.99 | degree |
| Vertical pipe weigh section, $w_v$ | 19.50 | 19.50 | lbf/ft |
| Vertical length | 4,207 | 4,208 | ft |
| Vertical force, $W_v$ | 82,049 | 82,057 | lbf |
| Axial compressive force at heel, $F_{pi/2}$ | 243,750 | 241,475 | lbf |
| Catenary force, $F_{ct}$ |  | 354,162 | lbf |
| Axial compressive force in string at KOP, $F_o$ | 1,270,187 | 1,010,575 | lbf |
| Tension at surface, T | 1,352,236 | 1,092,633 | lbf |

**Sensitivity Analysis:**

Two uncertain parameters can be varied. They are pipe weight and friction coefficient. The pipe weight of the curve section differs from 85 lbf/ft to 95 lbf/ft, while the friction coefficient of the horizontal section ranges from 1.5 to 2.5. The increase in friction coefficient would cause an increase in hook load. In fact, with the pipe weight = 85 lbf/ft and friction coefficient=1.5, the total hook load for catenary design is about 956,199 lbf while it is approximately 973,907 lbf with friction coefficient=1.6. Also, the total hook load would increase with the increase in pipe weight. The total hook load varies from 956,189 lbf to 1,02,7789 lbf in the range of pipe weight of the curve section from 85 to 95 lbf/ft. The total hook load plots with different friction coefficients and pipe weights are presented in Figures 8 to 18.

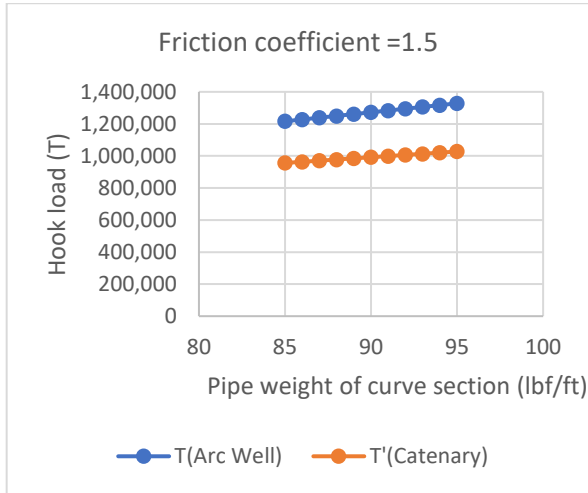

**Figure 8a**: Hook load profile with µ=1.5

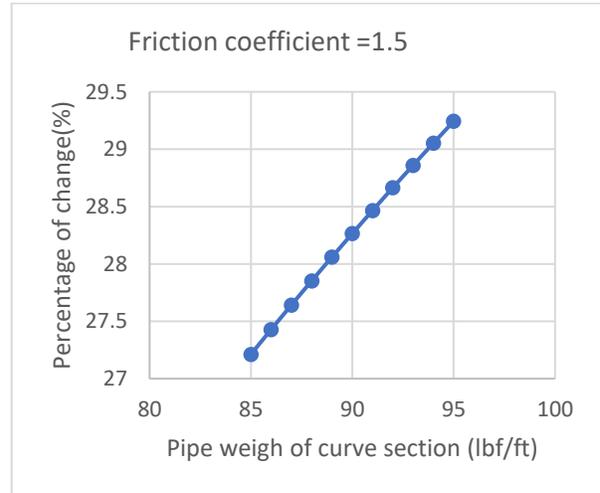

**Figure 8b**: Percentage of hook load difference

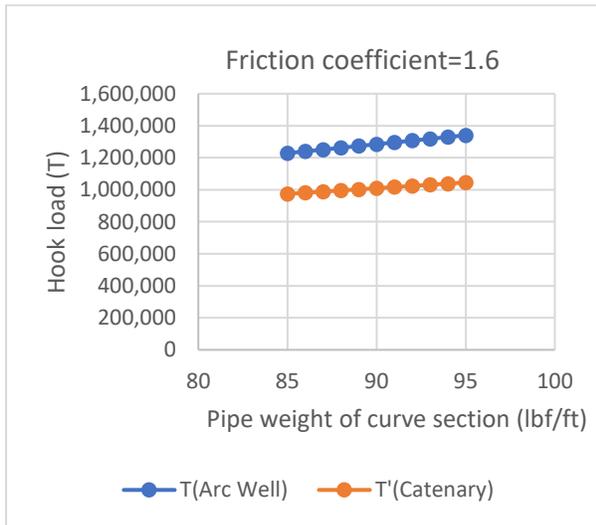

**Figure 9a**: Hook load profile with µ=1.6

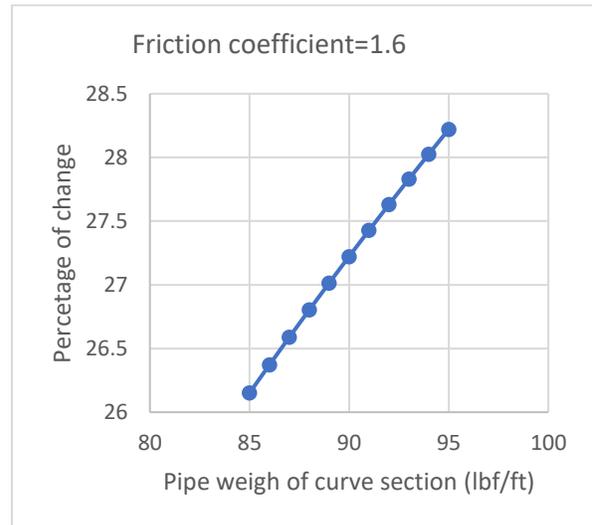

**Figure 8b**: Percentage of hook load difference

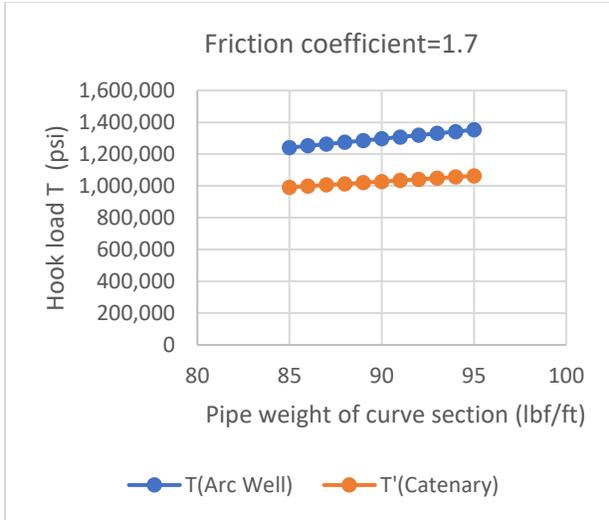 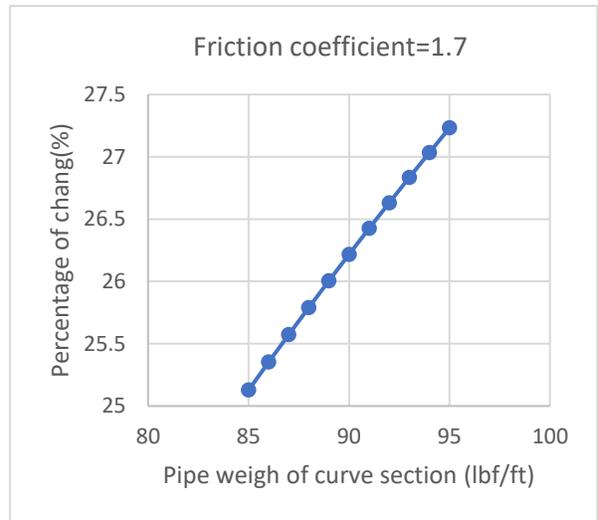

**Figure 10a**: Hook load profile with μ=1.7     **Figure 10b**: Percentage of hook load difference

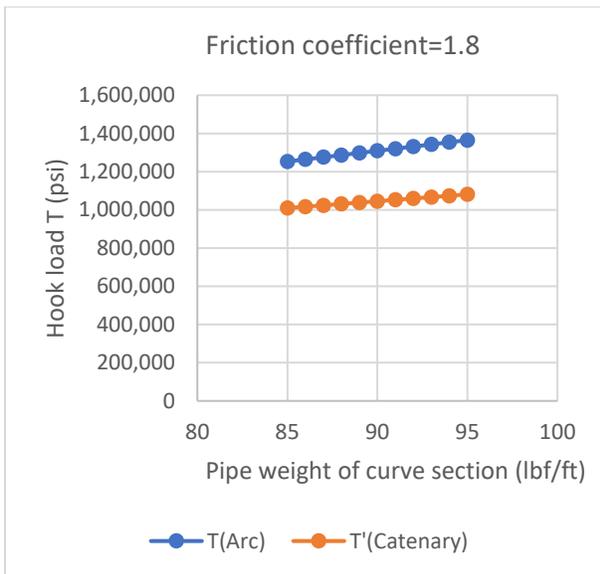 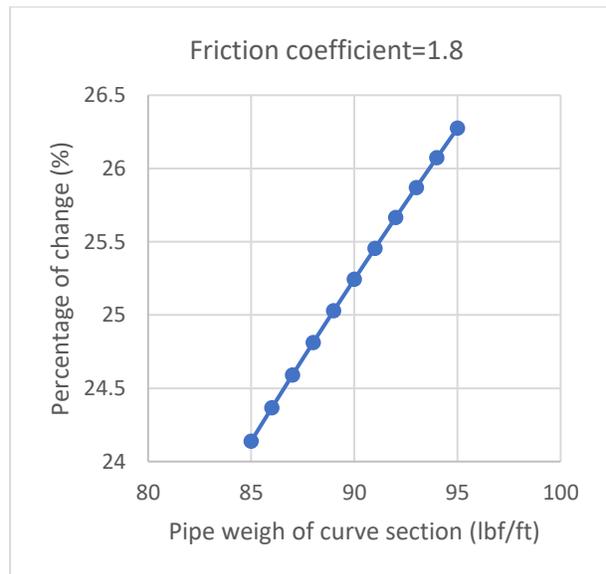

**Figure 11a**. Hook load profile with μ=1.8     **Figure 11b**: Percentage of hook load difference

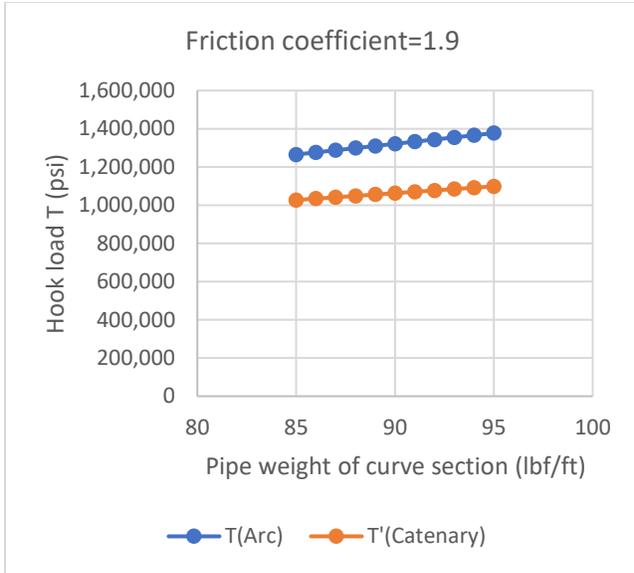

**Figure 12a**: Hook load profile with μ=1.9

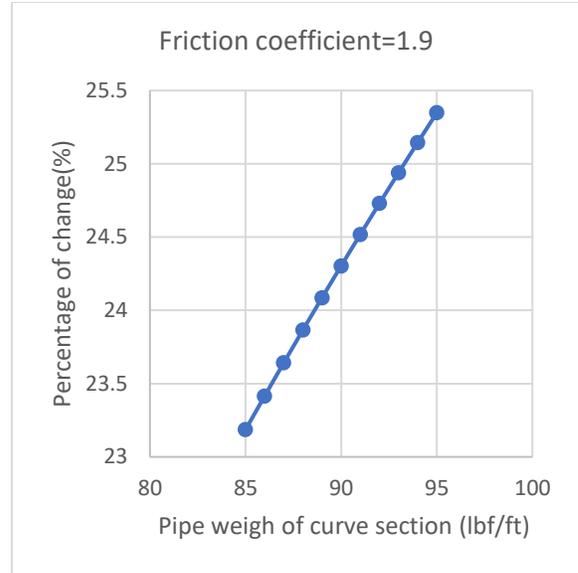

**Figure 12b**: Percentage of hook load difference

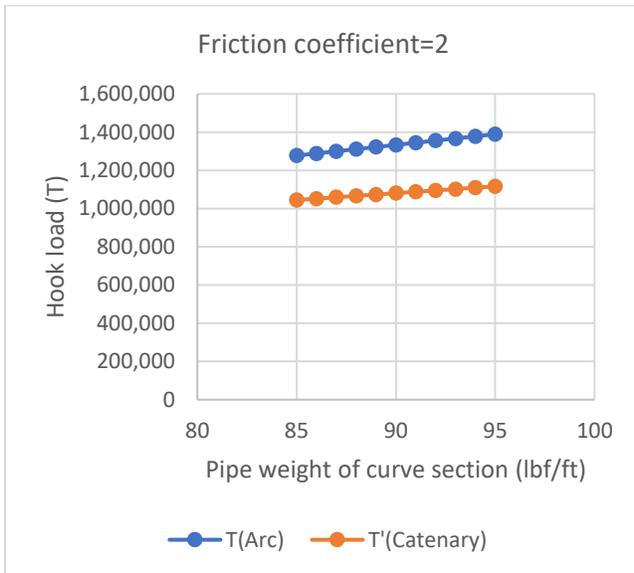

**Figure 13a**: Hook load profile with μ=2

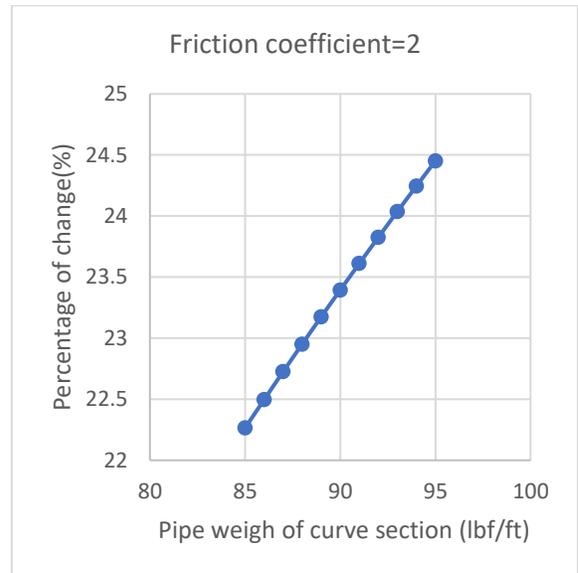

**Figure 13b**: Percentage of hook load difference

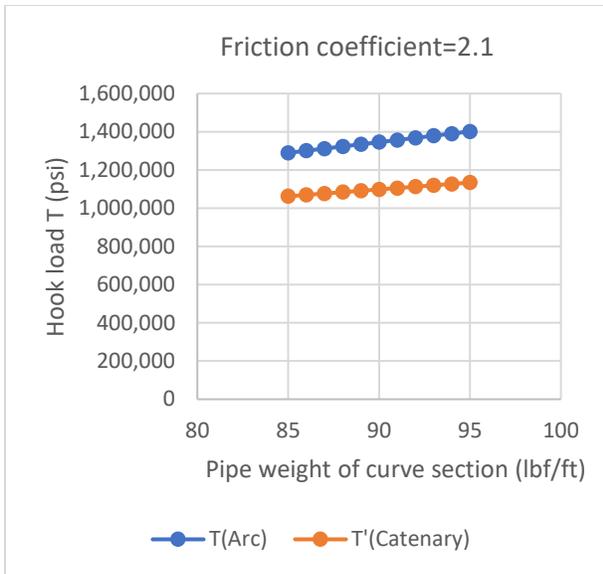

**Figure 14a**: Hook load profile with μ=2.1

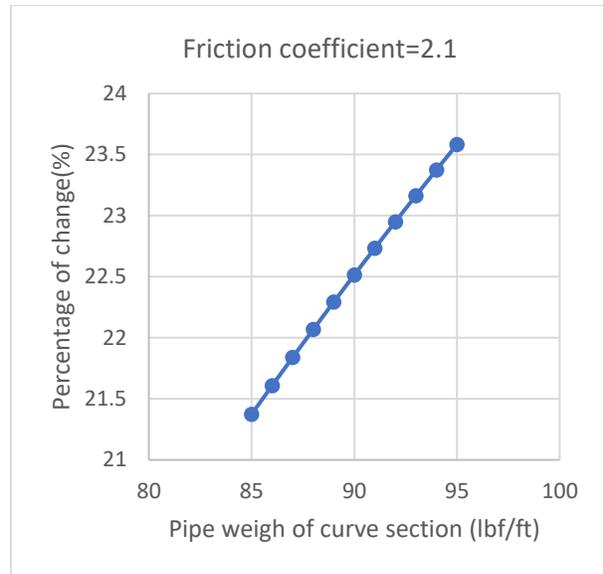

**Figure 14b**: Percentage of hook load difference

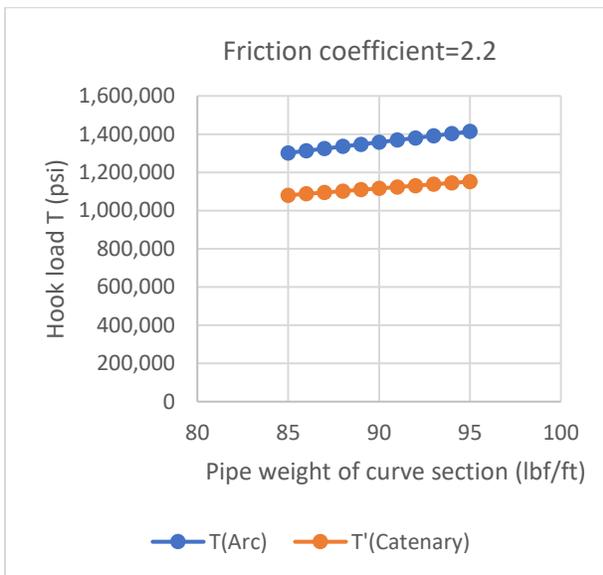

**Figure 15a**: Hook load profile with μ=2.2

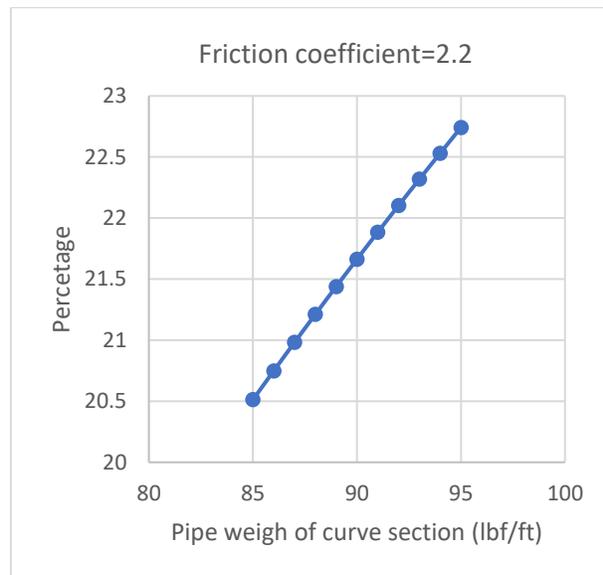

**Figure 15b**: Percentage of hook load difference

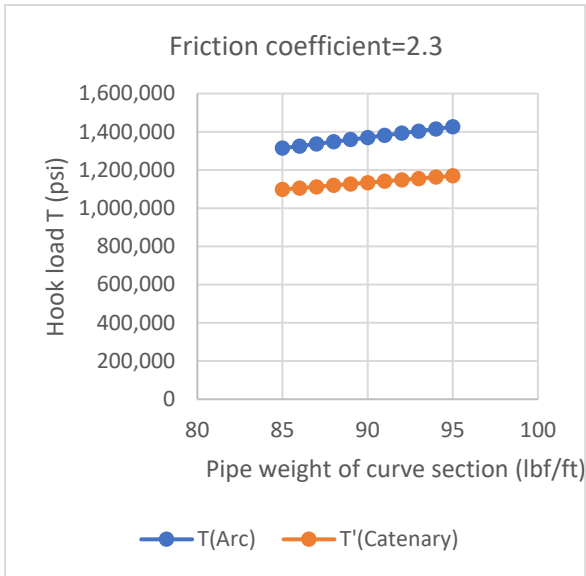 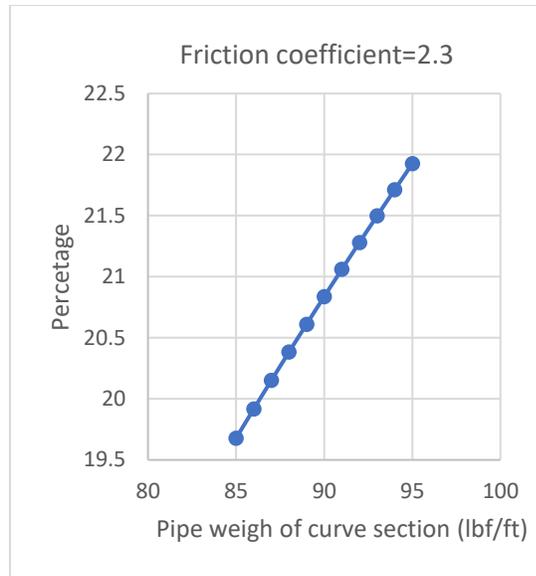

**Figure 16a**: Hook load profile with μ=2.3      **Figure 16b**: Percentage of hook load difference

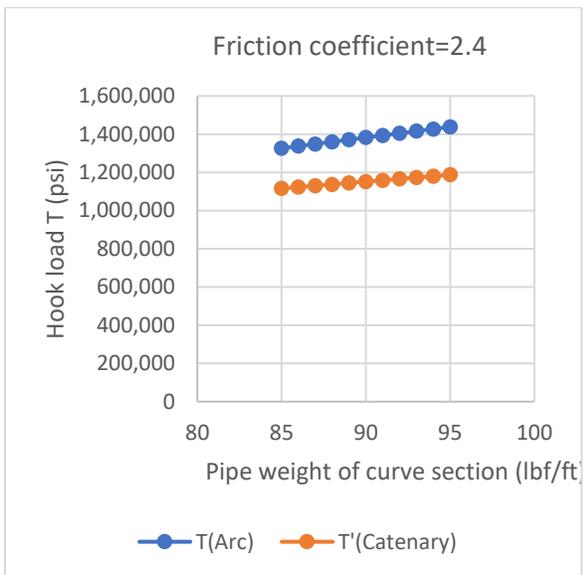 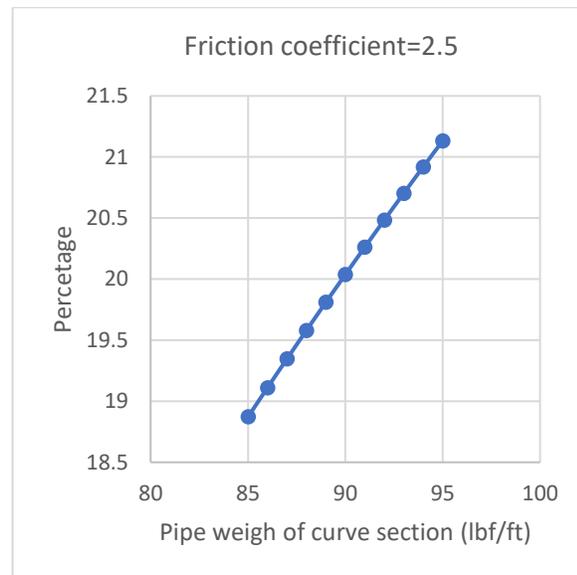

**Figure 17a**: Hook load profile with μ=2.4      **Figure 17b**: Percentage of hook load difference

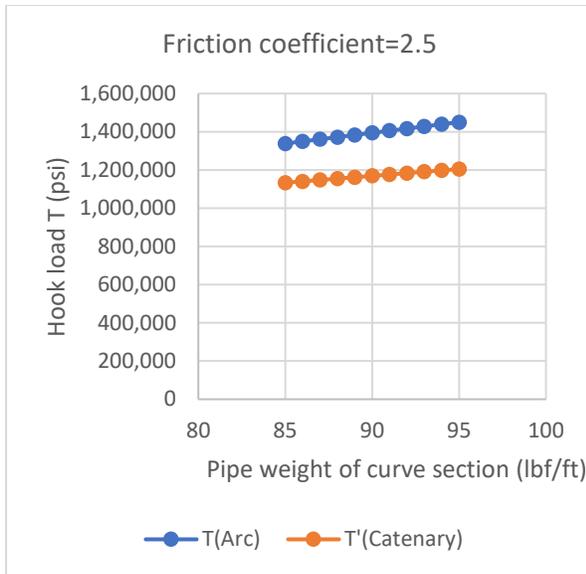 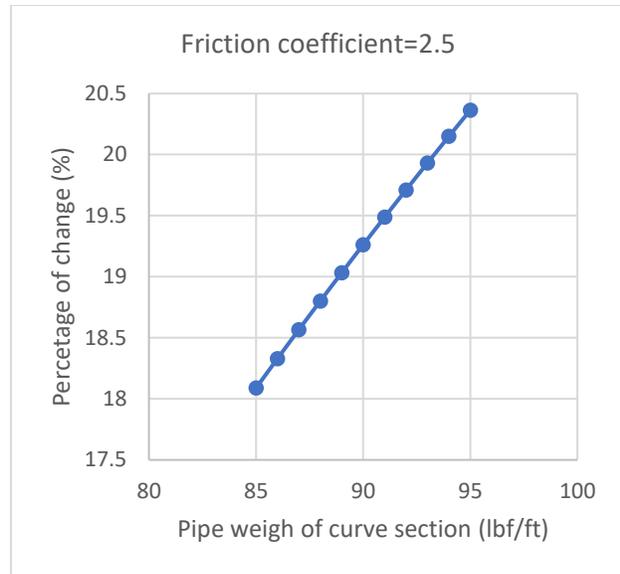

**Figure 18a**: Hook load profile with μ=2.5       **Figure 18b**: Percentage of hook load difference

## Discussion

Based on Figure 6, the two methods have a similar profile of North Displacement versus East Displacement because they have similar horizontal displacement. The difference in vertical displacement would create a distinction between the two methods.

The total hook load for arc trajectory design is 1,352,236 lbf which is far greater than the total hook load of catenary trajectory design (1,092,633 lbf). Based on the outcome from Table 4, the hook load difference is caused by friction in the curve section. In fact, the axial compressive force in the string at KOP for acr trajectory is 1,270,187 lbf compared to 1,010,575 lb for catenary trajectory design. The difference is 23.8 %.

## Conclusions

Several conclusions can be withdrawn from this study:
1. A mathematical model applied for this study is simple and straightforward. It uses closed-form equations and removes the complexity of numerical calculation.
2. The difference in vertical displacement would create a distinction between the two methods. This could cause the disparity of total hook load.
3. The outcome has shown that catenary trajectory design would reduce the total hook load compared to traditional arc design. The reduction is about 23.8 %.

## Acknowledgement

The author is very grateful for Dr. Guo's instruction to complete this paper.

## References

1. Ryan, G., Reynolds, J., and F. Raitt. "Advances in Extended Reach Drilling - An Eye to 10 km Stepout." Paper presented at the SPE Annual Technical Conference and Exhibition, Dallas, Texas, October 1995. doi: https://doi.org/10.2118/30451-MS
2. Meader, Tony, Allen, Frank, and Graham Riley. "To the Limit and Beyond - The Secret of World-Class Extended-Reach Drilling Performance at Wytch Farm." Paper presented at the IADC/SPE Drilling Conference, New Orleans, Louisiana, February 2000. doi: https://doi.org/10.2118/59204-MS
3. Mason, C.J., and A. Judzis. "Extended-Reach Drilling -- What is the Limit?." Paper presented at the SPE Annual Technical Conference and Exhibition, New Orleans, Louisiana, September 1998. doi: https://doi.org/10.2118/48943-MS
4. McClendon, R.T. "Directional Drilling Using the Catenary Method." Paper presented at the SPE/IADC Drilling Conference, New Orleans, Louisiana, March 1985. doi: https://doi.org/10.2118/13478-MS
5. Aadnoy, B.S., and Ketil Andersen. "Friction Analysis for Long-Reach Wells." Paper presented at the IADC/SPE Drilling Conference, Dallas, Texas, March 1998. doi: https://doi.org/10.2118/39391-MS
6. Bernt S Aadnøy, Ketil Andersen. "Design of oil wells using analytical friction models. Journal of Petroleum Science and Engineering." Volume 32, Issue 1, 2001, Pages 53-71, https://doi.org/10.1016/S0920-4105(01)00147-4.
7. Ma, Shanzhou, Huang, Genlu, Zhang, Jianguo, and Zhiyong Han. "Study on Design of Extended Reach Well Trajectory." Paper presented at the SPE International Oil and Gas Conference and Exhibition in China, Beijing, China, November 1998. doi: https://doi.org/10.2118/50900-MS
8. Planeix, Michele Y., and Richard C. Fox. "Use Of An Exact Mathematical Formulation To Plan Three Dimensional Directional Wells." Paper presented at the SPE Annual Technical Conference and Exhibition, Las Vegas, Nevada, September 1979. doi: https://doi.org/10.2118/8338-MS
9. Maidla, E.E., and J.H.B. Sampaio. "Field Verification of Lead Angle and Azimuth Rate of Change Predictions in Directional Wells Using a New Mathematical Model." Paper presented at the SPE Eastern Regional Meeting, Morgantown, West Virginia, October 1989. doi: https://doi.org/10.2118/SPE-19337-MS
10. Nguyen, V., Olayiwola, O., Guo, B., & Liu, N. (2021). Well Cement Degradation and Wellbore Integrity in Geological CO2 Storages: A Literature Review. *Petroleum & Petrochemical Engineering Journal, 5(3)*
11. Olayiwola, O., Nguyen, V., Liu, N., & Guo, B. (2023). Prior assessment of CO 2 leak rate through cracks sealed by nanoparticle gels. Journal of Petroleum Exploration & Production Technology, 13(6).
12. Carrera, M., Zarooni, M., Olayiwola, O., Nguyen, V., & Boukadi, F. (2023). Impacts of Asphaltene Deposition on Oil Recovery following a Waterflood–A Numerical Simulation Study. *Available at SSRN 4508842*.
13. Olayiwola, O., Nguyen, V., Yousuf, N., Baudoin, N., Liu, N., & Guo, B. (2022a). Experimental Investigation of Nanosilica Gel Properties for Well Integrity Remediation. Energy & Fuels, 36(23), 14094-14100.
14. Nguyen, V., Olayiwola, O., Liu, N., & Guo, B. (2022). Investigation of Nano-Silica Solution Flow through Cement Cracks. Sustainability, 15(1), 577.

**Appendix I**

The catenary curve in two Cartesian Coordinates is presented in Figure 19. A simple catenary curve can be expressed as:

$$y = \mathrm{acosh}\left(\frac{x}{a}\right) = \frac{a}{2}\left(e^{\frac{S-S_{end}}{a}} + e^{-\frac{S-S_{end}}{2}}\right) (24)$$

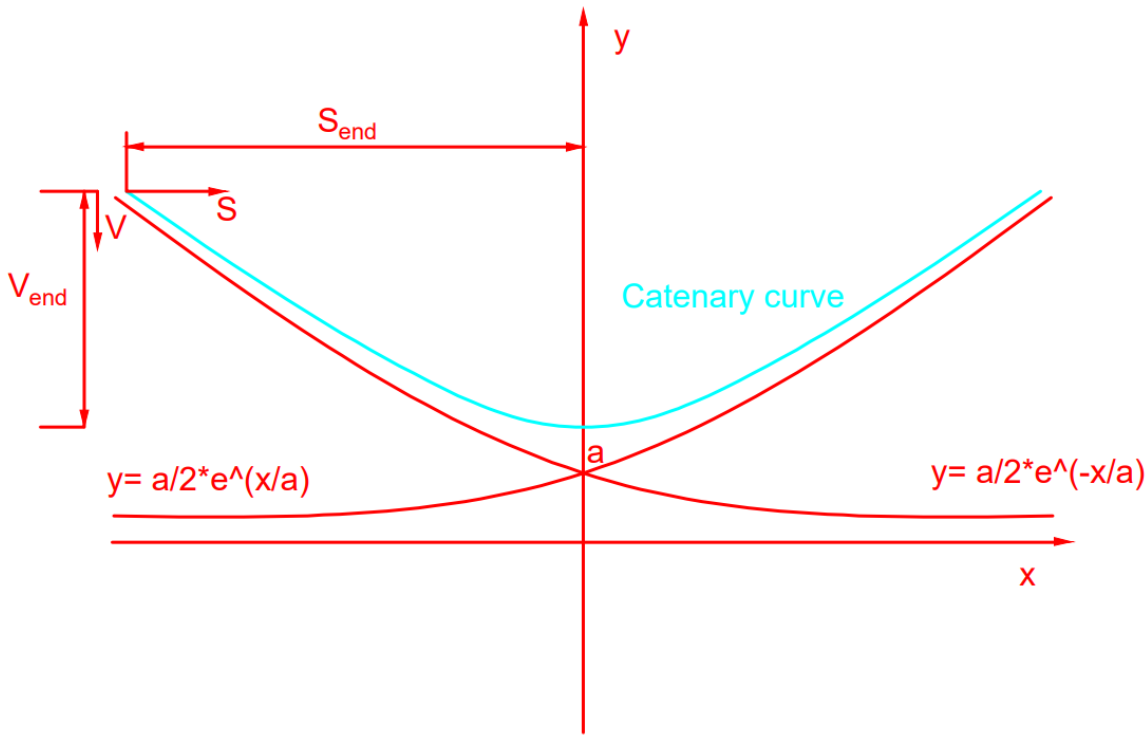

**Figure 19**: Catenary curve in 2D Cartesian Coordinate

The vertical displacement can be computed as:

$$V = V_{end} - (y - a) = V_{end} - \left\{\frac{a}{2}\left[e^{\frac{S-S_{end}}{a}} + e^{-\frac{S-S_{end}}{a}}\right] - a\right\} (25)$$

Where a can be found by solving the following equation:

$$V_{end} = \frac{a}{2}\left[e^{-\frac{S_{end}}{a}} + e^{\frac{S_{end}}{a}}\right] \quad (26)$$

The curvature of catenary is a function of horizontal displacement as below:

$$C = \frac{\frac{d^2V}{dS^2}}{\left(1 + \left(\frac{dV}{dS}\right)\right)^{\frac{3}{2}}} \quad (27)$$

The detail form of curvature can be express as:

$$C = \frac{-\frac{1}{2a}\left[e^{\frac{S-S_{end}}{a}} + e^{-\frac{S-S_{end}}{a}}\right]}{\left(1 + \left(-\frac{1}{2}\left[e^{\frac{S-S_{end}}{a}} - e^{-\frac{S-S_{end}}{a}}\right]\right)^2\right)^{\frac{3}{2}}} \quad (28)$$

The radius of curvature is:

$$R = \frac{1}{C} \quad (29)$$

The build rate can be calculated by:

$$B = 5730C \quad (30)$$

The inclination angle can be expressed as:

$$\theta = \frac{\pi}{2} - \alpha = \frac{\pi}{2} - arctan\frac{dV}{dS} = \frac{\pi}{2} - \arctan\left(-\frac{1}{2}[e^{\frac{S-S_{end}}{a}} - e^{-\frac{S-S_{end}}{a}}]\right) \quad (31)$$

**Appendix II**

**Arc design**

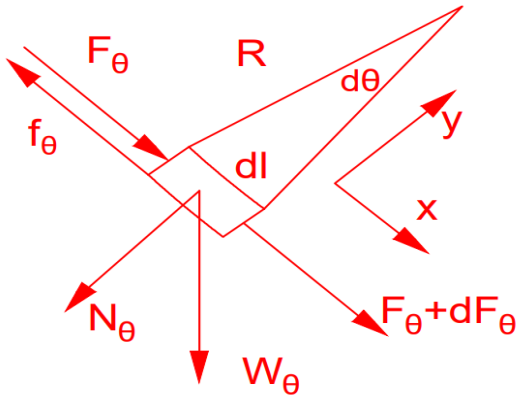

**Figure 20**: Free-body diagram of a portion of a rope-like string in a curve hole section

$$\sum F = 0 \quad (32)$$

In x direction:

$$dF_\theta = W_\theta \cos\theta - f_\theta \quad (33)$$

In y direction:

$$N = W_\theta \sin\theta \quad (34)$$

Also,

$$W_\theta = w_c R d\theta \quad (35)$$

$$f_\theta = \mu N = \mu W_\theta \sin\theta \quad (36)$$

$$dF_\theta = w_c R d\theta \cos\theta - \mu w_c R d\theta \sin\theta = w_c R d\theta [\cos\theta - \mu \sin\theta] \quad (37)$$

Integrate Eqn.37 to obtain:

$$\int_{F_{\theta_1}}^{F_{\theta_2}} dF_\theta = \int_{\theta_1}^{\theta_2} w_c R [\cos\theta - \mu \sin\theta] d\theta \quad (38)$$

$$F_{\theta_2} - F_{\theta_1} = w_c R [\sin\theta_2 + \mu\cos\theta_2 - \sin\theta_1 - \mu\cos\theta] \quad (39)$$

For $\theta_2 = \pi/2$, $\theta_1 = 0$

The axial compressive force at the heel can be expressed by the following equation:

$$F_{\pi/2} = \mu * W_h \quad (40)$$

The axial force in the curve section can be computed as:

$$F_O = F_{\frac{\pi}{2}} + w_c R(\mu + 1) \quad (41)$$

The Total hook load in Figure 7 is the combination of axial force in the curve section and Vertical force $W_v$ generated by pipe weight:

$$T = F_o + W_v \quad (42)$$

**Catenary design**

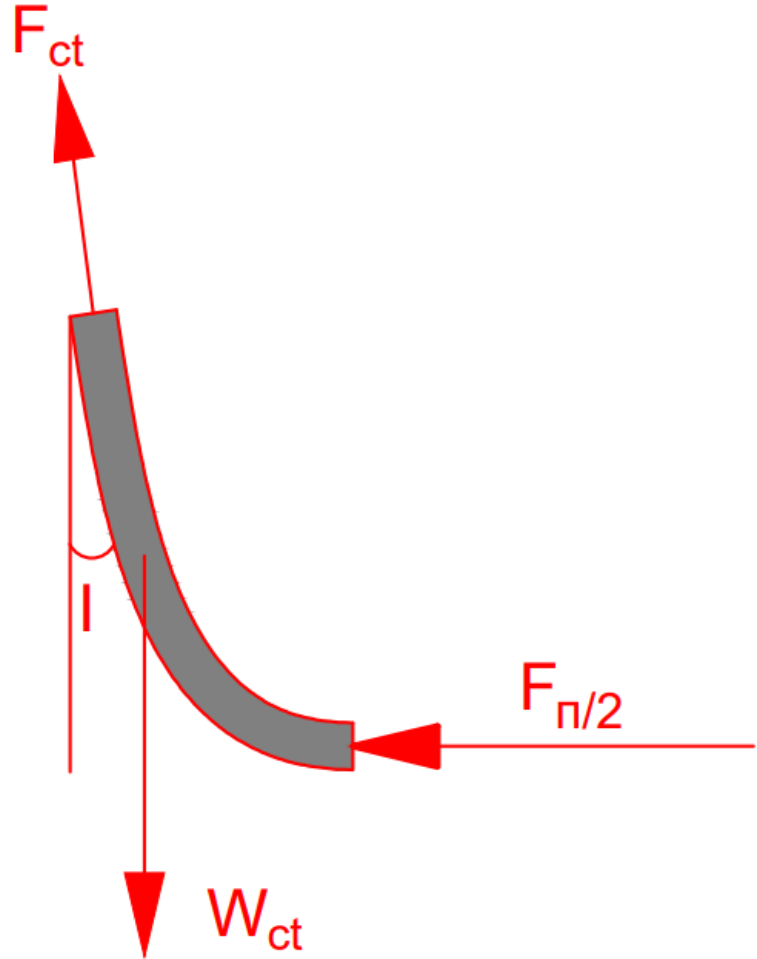

**Figure 20**: Force Analysis on catenary section

The axial compressive force at the heel has a similar expression with Arc design:

$$F_{\pi/2} = \mu * w_h \quad (43)$$

The influence of catenary shape on axial compressive force at the heel can be expressed as $F_{ct}$:

$$F_{ct} = F_{\frac{\pi}{2}}/\sin(I) \quad (44)$$

The axial force in the catenary can be computed as:

$$F_O = F_{ct} + w_c R[\sin(I) + \mu(1 - \cos(I))] \quad (45)$$

The total hook load is the combination of axial force in the curve section and Vertical force $W_v$ generated by pipe weight:

$$T = F_o + W_V \quad (46)$$